\documentclass{article}

\PassOptionsToPackage{numbers, compress}{natbib}
\usepackage[preprint]{neurips_2026}
\usepackage[utf8]{inputenc} 
\usepackage[T1]{fontenc}    
\usepackage{hyperref}       
\usepackage{url}            
\usepackage{booktabs}       
\usepackage{amsfonts}       
\usepackage{nicefrac}       
\usepackage{microtype}      
\usepackage{xcolor}         
\usepackage{amssymb} 
\usepackage{amsthm}
\usepackage{amsmath} 
\usepackage{needspace}
\usepackage{caption}
\usepackage{graphicx}
\usepackage{makecell}
\usepackage{array}
\usepackage{wrapfig}

\usepackage{multirow}
\usepackage[table]{xcolor}
\usepackage{array}
\usepackage{tabularx}
\usepackage{makecell}

\newcolumntype{Y}{>{\centering\arraybackslash}X}
\newcolumntype{C}[1]{>{\centering\arraybackslash}m{#1}}

\newcommand{\ourscell}[1]{\cellcolor{gray!30}#1}
\newcommand{\oursmethod}[1]{\cellcolor{gray!30}\textbf{#1}}

\usepackage{wrapfig}
\usepackage{graphicx}
\usepackage{array}

\newtheorem{proposition}{Proposition}
\title{TCP-SSM: Efficient Vision State Space Models with Token-Conditioned Poles}

\author{
Sara Shoouri \quad Morteza Tavakoli Taba \quad Hun-Seok Kim \\
University of Michigan \\
\texttt{\{sshoouri, tmorteza, hunseok\}@umich.edu}
}

\begin{document}

\maketitle

\begin{abstract}

State Space Models (SSMs) have emerged as a compelling alternative to attention models for long-range vision tasks, offering input-dependent recurrence with linear complexity. However, most efficient SSM variants reduce computation cost by modifying scan routes, resolutions, or traversal patterns, while largely leaving the recurrent dynamics implicit. Consequently, the model's state-dependent memory behavior is difficult to control, particularly in compact backbones where long scan paths can exceed the effective memory horizon. We propose Token-Conditioned Poles SSM (TCP-SSM), a structured selective SSM framework that improves efficiency while making recurrence dynamics explicit and interpretable through stable poles. TCP-SSM builds each scan operator with 1) real poles that model monotone or sign-alternating decay, and 2) complex-conjugate poles that capture damped oscillatory responses. Using bounded radius and angle modulation, TCP-SSM converts shared base poles into token-dependent poles, allowing each scan step to adapt its memory behavior to the current visual token while preserving pole stability. For practical scalability, we integrate grouped pole sharing with a lightweight low-rank input pathway, yielding an efficient scan operator that preserves linear-time scan complexity. Across image classification, semantic segmentation, and object detection, TCP-SSM reduces SSM computation complexity up to 44\% in Vision Mamba-style models while maintaining or surpassing baseline accuracy.

\end{abstract}

\vspace{-4mm}
\section{Introduction}
\vspace{-2mm}
Efficient long-range modeling remains a central challenge in modern computer vision. Convolutional Neural Networks \cite{krizhevsky2012imagenet, simonyan2014very, he2016deep, huang2019convolutional, liu2022convnet,chollet2017xception, szegedy2016rethinking} remain appealing due to their structured inductive biases and efficiency, but their localized kernels require long-range dependencies to be accumulated progressively through depth. Vision Transformers \cite{liu2021swin,bao2021beit, zamir2022restormer, dosovitskiy2020image, wang2021pyramid, touvron2021training,shoouri2026adaptive,shoouri2023efficient} address this limitation with self-attention, enabling direct global token interactions and strong long-range modeling, although their quadratic complexity scales poorly at high spatial resolutions. More recently, State Space Models (SSMs) \cite{gu2021efficiently, smith2022simplified, fu2022hungry, gu2023mamba, gu2021combining, gupta2022diagonal, liu2024vmamba, zhu2024vision}, especially Mamba-style selective scans, have emerged as compelling alternatives by combining linear-time sequence modeling with input-dependent selective dynamics. This balance between global context modeling and efficiency makes them well-suited for large-receptive-field vision tasks, including image classification, object detection, and segmentation.

Despite this promise, designing efficient visual SSMs remains challenging since visual data are inherently two-dimensional, whereas recurrence operates over a one-dimensional token ordering. Current Vision Mamba architectures address this mismatch with multi-directional cross-scanning strategies \cite{liu2024vmamba, zhu2024vision} that better preserve spatial structure after flattening. However, in compact backbones with limited hidden-state capacity, these trajectories can exceed the layer's effective memory horizon, causing distant information to attenuate before contributing meaningfully to the representation. Prior work addresses this through scan engineering, including skip sampling and multi-scale traversal schemes that reduce long-range forgetting \cite{pei2025efficientvmamba, yang2024plainmamba, huang2024localmamba, shi2024multi}. Although effective, these methods modify how spatial tokens are visited rather than how the recurrent dynamics store, retain, or propagate information. As a result, the scan's memory behavior remains largely implicit, making it difficult to characterize the learned timescales, signal decay rates, and adaptation to local image content.
\begin{figure}[t]
    \centering
    \includegraphics[width=1\textwidth]{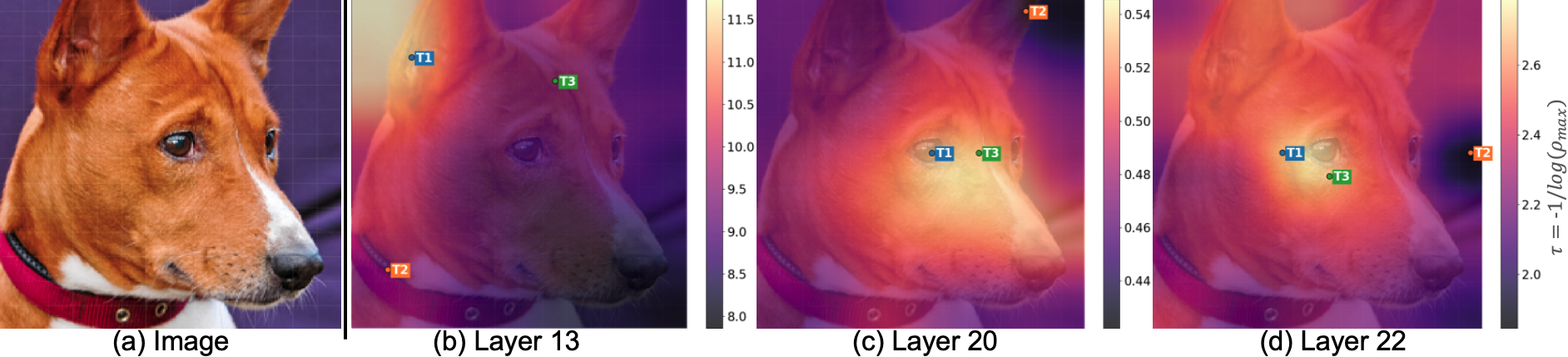}
    \caption{Illustration of the dominant pole-derived memory horizon at layers 13, 20, and 22 (of 24) for an example image. Brighter regions denote larger $\tau$ (slower decay, longer memory). T1, T2, and T3 highlight the longest memory, fastest decay, and strongest radius-weighted oscillation ($\rho_{\max}\theta/\pi$). Across depth, the pole-conditioned recurrence evolves from coarse object-level memory toward discriminative regions such as the face and eyes.
    }
    \label{fig:teaser}
    \vspace{-5mm}
\end{figure}

We take a different perspective and argue that efficient visual SSMs should be improved not only through scan design, but also through the recurrence structure itself. From a control theory perspective, the memory and frequency response of recurrent systems are governed by \textit{poles} in the complex plane, where the pole radius controls information decay and the pole angle determines localized oscillatory behavior \cite{maestre2017constrained, beck2022iterative}. Rather than relying on an unconstrained selective recurrence to implicitly discover suitable timescales, we introduce efficient SSMs with Token-Conditioned Poles (TCP-SSM) with an explicit and stable parameterization that adapts in a controlled manner. TCP-SSM equips the scan operator with a stable base denominator built from real and complex pole components, where real poles capture monotone memory decay and complex poles model oscillatory and frequency-selective behavior. The pole parameters vary with the input within bounded ranges, enabling spatially adaptive decay and oscillatory modes while preserving a compact and interpretable dynamical structure.

To ensure efficiency, we pair the structured pole parameterization with a lightweight low-rank input pathway, preserving SSM content selectivity without substantial hidden-state overhead. We further reduce computation through grouped pole sharing, allowing different channel groups to learn distinct dynamical profiles while avoiding fully independent poles for every feature dimension. This design retains the linear-time complexity of selective scanning while enhancing long-range expressiveness. In addition to improving efficiency, the pole-based formulation provides a direct view of the learned memory dynamics, which are often entangled within opaque state updates in conventional selective scans. Pole radii reveal how quickly past information is forgotten, pole angles characterize oscillatory modes and frequency preferences, and token-conditioned modulation shows how these behaviors vary across spatial contexts, offering a concrete interpretation of when the model preserves or suppresses information, as shown in Figure~\ref{fig:teaser}. We evaluate the proposed framework on Vision Mamba (Vim) backbones and EfficientVMamba (EVM) \cite{pei2025efficientvmamba}. Across image classification, segmentation and object detection, our method improves the accuracy-efficiency trade-off, reducing SSM computation by up to roughly 44\% in Vim and 70\% in EVM-Base while maintaining comparable/better accuracy.

Our contributions can be summarized as follows:
\vspace{-2mm}
\begin{itemize}
    \item We introduce a new structured selective SSM formulation with token-conditioned poles in which stable real and complex poles define the base recurrence dynamics, while token-conditioned modulation adapts memory and frequency response to the input.
    \vspace{-1mm}
    \item We combine grouped pole sharing with a lightweight low-rank input pathway, preserving linear-time scanning while reducing long-range modeling complexity in visual SSMs.
    \vspace{-1mm}
    \item Across classification, segmentation, and object detection, explicit TCP-SSM improves the accuracy-efficiency trade-off of visual SSMs, reducing SSM computation by 44\% in Vim and 70\% in EVM-Base while maintaining comparable/improved accuracy.
    \vspace{-2mm}
\end{itemize}
\vspace{-3mm}
\section{Related work}
\vspace{-2mm}
\subsection{Efficient Vision Backbones}
\vspace{-2mm}
Computationally efficient vision backbones have evolved from lightweight convolutional networks to mobile-oriented transformers and hybrid architectures. Early designs such as ShuffleNetV2 \cite{ma2018shufflenet} improve practical efficiency by optimizing memory access costs, while MobileNetV3 \cite{koonce2021mobilenetv3} advances accuracy-latency trade-offs through hardware-aware neural architecture search and platform-specific refinement. GhostNetV2 \cite{tang2022ghostnetv2} further improves inexpensive convolutional feature generation with hardware-friendly long-range spatial interaction. Hardware-aware search frameworks such as Once-for-All \cite{cai2019once} and FBNetV3 \cite{dai2021fbnetv3} extend this direction by learning adaptable model families or jointly optimizing architectures and training recipes under deployment constraints. These ideas have also shaped transformer and hybrid backbones, where MobileViT \cite{mehta2021mobilevit} introduces global token mixing into mobile models, MobileViTv2 \cite{mehta2022separable} reduces latency with separable linear-complexity attention, and AutoFormer \cite{wu2021autoformer}, EfficientFormerV2 \cite{li2023rethinking}, EfficientViT \cite{liu2023efficientvit}, and FasterViT \cite{hatamizadeh2023fastervit} improve mobile transformer efficiency through search, memory-efficient attention, or hierarchical design. Recent hybrids such as SwiftFormer \cite{shaker2023swiftformer}, RepViT \cite{wang2024repvit}, and MobileNetV4 \cite{qin2024mobilenetv4} further explore additive attention, transformer-inspired convolutions, and unified search spaces. Our work is complementary to this line of research, but instead of introducing another lightweight backbone or attention mechanism, we improve the recurrence dynamics of compact vision state space models.
\vspace{-3mm}
\subsection{State Space Models for Visual Representation}
\vspace{-3mm}
State space models have become competitive alternatives to attention for visual representation learning, with S4 \cite{gu2021efficiently}, S5 \cite{smith2022simplified}, Mamba \cite{gu2023mamba}, and Mamba-2 \cite{dao2024transformers} establishing efficient long-range modeling through structured state space parameterizations, input-dependent selective dynamics, and state space duality. In vision, S4ND \cite{nguyen2022s4nd} extends SSMs to images and videos, while Vision Mamba \cite{zhu2024vision} and VMamba \cite{liu2024vmamba} demonstrate strong generic backbones through bidirectional modeling and 2D cross-scan operators. This paradigm has expanded through Mamba-ND \cite{li2024mamba} for multidimensional data, MambaVision \cite{hatamizadeh2025mambavision} for hybrid Mamba-Transformer backbones, Mamba-R \cite{wang2025mamba} for register-token scaling, and V2M \cite{wang2024v2m} and 2DMamba \cite{zhang20252dmamba} for native two-dimensional state updates. VM-UNet \cite{ruan2024vm}, SegMamba \cite{xing2025segmamba}, and PointMamba \cite{liang2024pointmamba} further show their versatility across visual domains. To improve 2D efficiency, LocalMamba \cite{huang2024localmamba}, PlainMamba \cite{yang2024plainmamba}, and EfficientVMamba \cite{pei2025efficientvmamba} optimize spatial token serialization through windowed, continuous, or atrous scanning, while DefMamba \cite{liu2025defmamba} and Multi-Scale VMamba \cite{shi2024multi} introduce adaptive and multi-scale traversals. Other works redesign the state space block itself, including VSSD \cite{shi2025vssd} with non-causal state space duality, EfficientViM \cite{lee2025efficientvim} with compressed hidden states, and GroupMamba \cite{shaker2025groupmamba} with grouped visual state space modeling. Our work shares the goal of efficient visual SSM design but directly structures recurrence through stable poles and token-conditioned modulation, enabling efficient and interpretable long-range modeling.
\vspace{-3mm}
\section{Method}
\label{Main:Method}
\vspace{-3mm}
\setlength{\abovedisplayskip}{2pt}
\setlength{\belowdisplayskip}{2pt}
\setlength{\abovedisplayshortskip}{2pt}
\setlength{\belowdisplayshortskip}{2pt}
\begin{figure}[t]
    \centering
    \includegraphics[width=\textwidth]{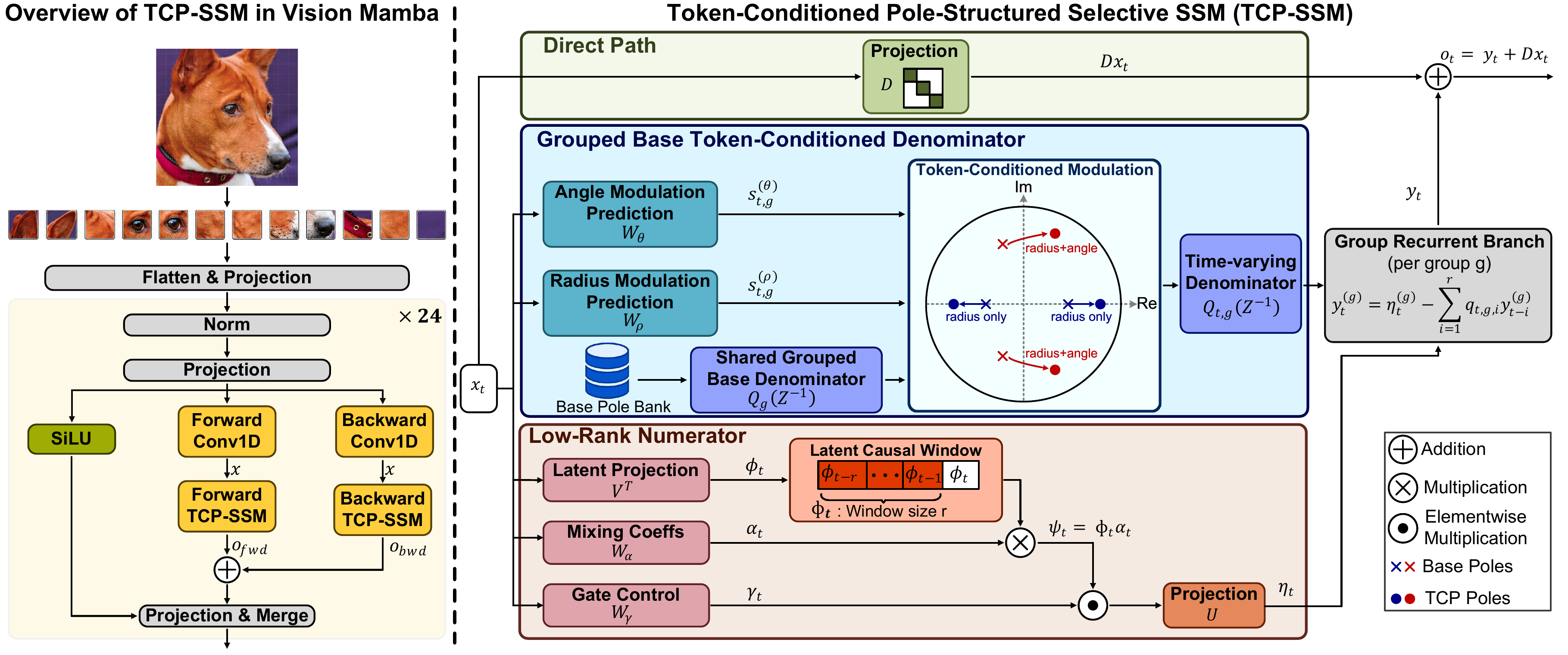}
    \caption{Overview of TCP-SSM Model. Left: TCP-SSM is inserted into the Vision Mamba model. Right: For token $x_t$, a shared grouped pole bank defines a stable base denominator $Q_g(z^{-1})$, while token-conditioned radius and angle modulation produce the local denominator $Q_{t,g}(z^{-1})$.
A lightweight low-rank numerator forms the driving signal $\eta_t$, and the grouped recurrent branch outputs $y_t$. The final output $o_t$ combines recurrent memory with a learned diagonal direct path.
    }
    \label{fig:Main_overview}
    \vspace{-5mm}
\end{figure}
This section introduces the proposed TCP-SSM: \emph{Efficient State Space Model with Token-Conditioned Poles}. The overview of TCP-SSM is demonstrated in Figure \ref{fig:Main_overview}. We first establish notation, review state space and control theory preliminaries, and then present our algorithm.
\vspace{-2mm}
\paragraph{Notation:} Let $F \in \mathbb{R}^{B \times H \times W \times E}$ be the input feature map with batch size $B$, spatial dimensions of $H \times W$, and $E$ channels. A scan rule flattens $F$ into a token sequence $X \in \mathbb{R}^{B \times M \times E}$, where $M = HW$. For clarity, we describe the method for a single scan route. In practice, it is applied independently along each scan direction of a Vision Mamba-style backbone, with outputs fused exactly as in the original architecture.

We partition the $E$ channels into $G$ equal groups of size $E_g = E / G$. For a token $x_t \in \mathbb{R}^{E}$ at position $t$, its $g$-th group is $x_t^{(g)} \in \mathbb{R}^{E_g}$. The recurrent output prior to the direct path is $y_t \in \mathbb{R}^E$, with its group block denoted $y_t^{(g)} \in \mathbb{R}^{E_g}$. The final processed output is $o_t \in \mathbb{R}^E$. We define $r$ as the denominator polynomial order and $r_f$ as the low-rank numerator dimension. We reserve the scalar variables $u_t$ and $v_t$ for classical dynamical systems and use $x_t$, $y_t$, and $o_t$ for multidimensional feature representations in our model.
\vspace{-3mm}
\subsection{Preliminaries}
\vspace{-2mm}
\paragraph{State Space Models:} Classical continuous-time SSMs map a scalar input $u(\tau) \in \mathbb{R}$ to an output $v(\tau) \in \mathbb{R}$ via a latent state $h(\tau) \in \mathbb{R}^{N}$:
\begin{equation}
\dot{h}(\tau) = A_c h(\tau) + b_c u(\tau),
\qquad
v(\tau) = c^\top h(\tau) + d\,u(\tau),
\label{eq:continuous_ssm}
\end{equation}

where $A_c \in \mathbb{R}^{N \times N}$ is the continuous transition matrix, $b_c \in \mathbb{R}^{N}$ injects the input into the latent state, $c \in \mathbb{R}^{N}$ maps the state to the output, and $d \in \mathbb{R}$ is a direct feedthrough term. Here $N$ denotes the latent dynamical system size.

Discretizing this system with a step size $\Delta > 0$ under a zero-order hold (ZOH) \cite{wang2008zoh} assumption yields the discrete-time update:
\begin{equation}
h_t = A h_{t-1} + b\,u_t,
\qquad
v_t = c^\top h_t + d\,u_t.
\label{eq:discrete_ssm}
\end{equation}
When $A_c$ is invertible, then $A = \exp(\Delta A_c)$ and $b = A_c^{-1}\bigl(\exp(\Delta A_c) - I\bigr)b_c$, where $I$ is the identity matrix.

In visual backbones, the input at position $t$ is a feature token $x_t \in \mathbb{R}^E$. Thus, the scalar state space model is replicated across all $E$ channels. For a given channel $e$, the scalar input $x_{t,e}$ drives a distinct latent state $h_{t,e} \in \mathbb{R}^N$ according to $h_{t,e} = A_e h_{t-1,e} + b_e x_{t,e}$ and $v_{t,e} = c_e^\top h_{t,e} + d_e x_{t,e}$. Selective SSMs enhance this baseline by making the discretization and state interactions token-dependent, dynamically adapting memory at each spatial position.
\vspace{-3mm}
\paragraph{Control-Theoretic View:} A control-theoretic perspective in the $z$-domain reveals the memory mechanism more explicitly.  Consider the discrete-time linear time-invariant system (LTI) from Eq. \ref{eq:discrete_ssm}.
Assuming a zero initial state, the transfer function relating the input $U(z)$ and output $V(z)$ after the $z$-domain transform \cite{rantzer2011distributed} is defined as
\begingroup
\setlength{\abovedisplayskip}{1pt}
\setlength{\abovedisplayshortskip}{1pt}
\setlength{\belowdisplayshortskip}{1pt}
\setlength{\belowdisplayshortskip}{1pt}
\begin{equation}
H(z) \triangleq \frac{V(z)}{U(z)} = d + c^\top (I - A z^{-1})^{-1} b.
\label{eq:transfer_function}
\end{equation}
\endgroup
This function can equivalently be expressed in rational form as $H(z) =\frac{P(z^{-1})}{Q(z^{-1})}$, where $P$ and $Q$ are the numerator and denominator polynomials. A denominator of order $r$ expands to $Q(z^{-1}) = 1 + \sum_{i=1}^{r} q_i z^{-i}$. Assuming a unit numerator, the corresponding difference equation $v_t = u_t - \sum_{i=1}^{r} q_i\, v_{t-i}$ demonstrates that the denominator uniquely dictates how past information persists or decays. These dynamics are governed strictly by the polynomial roots (i.e., the system poles), which equal the eigenvalues of $A$ (see Section \ref{supp:transfer_function_derivation}). System stability requires all poles to reside strictly inside the unit disk \cite{van2002identification}.

Translating these insights into a practical architecture relies on two elementary stable factors in $Q(z^{-1})$. A first-order real factor $Q_r(z^{-1}) = 1 - a z^{-1}$ (where $|a| < 1$) enforces monotone ($a > 0$) or alternating ($a < 0$) exponential decay proportional to $a^k$, where $k$ is the discrete lag index. A second-order complex factor is defined as
\begingroup
\setlength{\abovedisplayskip}{2pt}
\setlength{\abovedisplayshortskip}{2pt}
\setlength{\belowdisplayshortskip}{1pt}
\setlength{\belowdisplayshortskip}{1pt}
\begin{equation}
Q_c(z^{-1}) = 1 - 2\rho \cos(\theta)\, z^{-1} + \rho^2 z^{-2},
\label{eq:complex_factor}
\end{equation}
\endgroup

where $0 < \rho < 1$ and $0 < \theta < \pi$. The corresponding conjugate poles $\rho e^{\pm j\theta}$ generate a decaying oscillation. The radius $\rho$ sets the exponential envelope and the angle $\theta$ determines the angular frequency, providing an interpretable basis for parameterizing recurrence denominators.

\vspace{-3mm}
\subsection{TCP-SSM: Token-Conditioned Pole-Structured Selective SSM}
\label{Main:TCP-SSM}
\vspace{-3mm}
The proposed formulation relies on three foundational components: 1) a structured denominator of stable real and complex-conjugate poles shared across channel groups, 2) data-dependent modulation of these poles, and 3) a lightweight low-rank numerator for the driving signal. This design preserves the linear-time efficiency of selective scanning while rendering the dynamics more interpretable, structured, and analytically tractable. All proofs are deferred to the supplementary material.

\vspace{-3mm}
\paragraph{Stable Grouped Base Denominator:} Each of $G$ channel groups shares a \emph{base denominator polynomial}, defining its pole bank prior to token-conditioned modulation. Given $L$ real and $K$ complex pole families (yielding a total order of $r = L + 2K$), the base denominator for group $g$ is:
\begingroup
\setlength{\abovedisplayskip}{1pt}
\setlength{\abovedisplayshortskip}{1pt}
\setlength{\belowdisplayshortskip}{1pt}
\setlength{\belowdisplayshortskip}{1pt}
\begin{equation}
Q_g(z^{-1}) =
\prod_{\ell=1}^{L} \left(1 - a_{g,\ell} z^{-1}\right)
\prod_{k=1}^{K} \left(1 - 2\rho_{g,k}\cos(\theta_{g,k}) z^{-1} + \rho_{g,k}^2 z^{-2}\right).
\label{eq:Qg_factorized}
\end{equation}
\endgroup

To guarantee stability, all poles must strictly reside inside the unit disk ($a_{g,\ell} \in (-1,1)$, $\rho_{g,k} \in (0,1)$, and $\theta_{g,k} \in (0,\pi)$). We enforce this by mapping unconstrained learned variables into valid ranges using a constant boundary margin $0<\varepsilon<1$. Complex poles are bounded via sigmoid function ($\sigma$) applied to trainable pole parameters $\hat{\rho}_{g,k}$ and $\hat{\theta}_{g,k}$  such that:
\begin{equation}
\rho_{g,k} = (1-\varepsilon)\sigma(\hat{\rho}_{g,k}),
\qquad
\theta_{g,k} = \pi \sigma(\hat{\theta}_{g,k}).
\label{eq:pole_parameterization}
\end{equation}
For real poles, we use a decoupled signed-magnitude parameterization that separates the response sign $\bar{s}_{g,\ell}$ from the magnitude $\bar{\rho}^{R}_{g,\ell}$, enabling the subsequent selective construction. Given trainable parameters $\hat{\rho}^{R}_{g,\ell}$ and $\hat{s}_{g,\ell}$, we define the real pole in a way that guarantees $|a_{g,\ell}| < 1$:
\begin{equation}
a_{g,\ell} = \bar{s}_{g,\ell}\,\bar{\rho}^{R}_{g,\ell},
\qquad
\bar{\rho}^{R}_{g,\ell} = (1-\varepsilon)\sigma(\hat{\rho}^{R}_{g,\ell}) \in (0,1),
\qquad
\bar{s}_{g,\ell} = \tanh(\hat{s}_{g,\ell}) \in (-1,1).
\label{eq:real_pole_param}
\end{equation}
\begin{proposition}
If $|a_{g,\ell}| < 1$ for all real factors and $0 < \rho_{g,k} < 1$, $0 < \theta_{g,k} < \pi$ for all complex factors, then all poles associated with $Q_g$ lie strictly inside the unit disk. Hence, each grouped base denominator is Schur stable \cite{borcea2009lee} (see Section \ref{supp:pole_stability_proofs}).
\end{proposition}
\vspace{-3mm}
\paragraph{Token-conditioned Grouped Denominator:} We introduce time-varying dynamics by replacing the base grouped denominator $Q_g$ with a token-dependent $Q_{t,g}$. The current token $x_t$ modulates the shared base poles to achieve input-dependent selectivity. For each group $g$, we obtain a strictly positive radius scale $s_{t,g}^{(\rho)}$ and a bounded angle scale $s_{t,g}^{(\theta)}$ via:
\begin{equation}
s_{t,g}^{(\rho)} =
\frac{\delta_{\min} + \operatorname{softplus}\!\left([W_\rho x_t + b_\rho]_{\kappa(g)}\right)}{\delta_0},
\qquad
s_{t,g}^{(\theta)} =
1 + \lambda_\theta \tanh\!\left([W_\theta x_t + b_\theta]_{\kappa(g)}\right).
\label{eq:token_modulation_scales}
\end{equation}
Here $\delta_{\min}\geq 0$, $\delta_0>0$, and $\lambda_\theta\in[0,1)$ are fixed hyperparameters, where $\delta_{\min}$ and $\delta_0$ set the offset and normalization of the radius scale and $\lambda_\theta$ bounds the angle modulation around one. The trainable parameters are $W_\rho,W_\theta \in \mathbb{R}^{C \times E}$ and $b_\rho,b_\theta \in \mathbb{R}^{C}$, where $C=G$ gives group-specific modulation and $C=1$ gives shared modulation. The index map satisfies $\kappa(g)=g$ in the group-specific case and $\kappa(g)=1$ in the shared case, so the obtained scales are broadcast across all groups. We adopt the shared configuration to improve the trade-off between performance and efficiency (see Table \ref{tab:pole_selectivity_ablation}). The $s_{t,g}^{(\rho)} > 0$ modulates the effective memory timescale by contracting or stretching pole magnitudes. The $s_{t,g}^{(\theta)}$, centered at 1, governs oscillatory behavior and its bounded $\tanh$ formulation prevents abrupt frequency shifts. These scales dynamically update the pole families as follows:

\textbf{1. Real Poles:} We update the signed-magnitude form as $a_{t,g,\ell} = \bar{s}_{g,\ell}\exp(s_{t,g}^{(\rho)} \log \bar{\rho}^{R}_{g,\ell})$. This scales the magnitude while preserving the sign $\bar{s}_{g,\ell}$, ensuring the response type (monotone vs. sign-alternating) remains fixed. Since  $s_{t,g}^{(\rho)} > 0$ and $\bar{\rho}^{R}_{g,\ell} \in (0,1)$, the modulated magnitude strictly remains in $(0,1)$, ensuring the pole stays safely inside the unit disk.

\textbf{2. Complex Poles:} We similarly adjust the decay envelope and oscillation frequency via:
\begingroup
\setlength{\abovedisplayskip}{2pt}
\setlength{\abovedisplayshortskip}{2pt}
\begin{equation}
\rho_{t,g,k}
=
\exp\!\left(s_{t,g}^{(\rho)} \log \rho_{g,k}\right),
\qquad
\theta_{t,g,k}
=
\operatorname{clip}\!\left(s_{t,g}^{(\theta)} \theta_{g,k}, 0, \pi\right).
\label{eq:token_conditioned_complex_params}
\end{equation}
\endgroup
This construction natively preserves $\rho_{t,g,k} \in (0,1)$ and enforces valid angular intervals, ensuring the poles remain in unit disk. Together with the real-pole update above, these modulations convert the stable base pole bank defined by $Q_g$ into a token-conditioned pole bank for scan step $t$. Specifically, the base poles $a_{g,\ell}$ and $\rho_{g,k}e^{\pm j\theta_{g,k}}$ are transformed into token-conditioned poles as $a_{t,g,\ell}$ and $\rho_{t,g,k}e^{\pm j\theta_{t,g,k}}$ using the modulation scales ($s_{t,g}^{(\rho /\ \theta )}$) predicted from $x_t$. These modulated poles define the denominator $Q_{t,g}$ used by the recurrent scan. When the modulation is identity, with $s_{t,g}^{(\rho /\ \theta )}=1$, the construction recovers the base denominator $Q_g$. 

Substituting these modulated poles back into the grouped base form yields the fully expanded token-conditioned denominator:
\begin{equation}
Q_{t,g}(z^{-1})
=
\prod_{\ell=1}^{L} \left(1-a_{t,g,\ell}z^{-1}\right)
\prod_{k=1}^{K} \left(1-2\rho_{t,g,k}\cos(\theta_{t,g,k})z^{-1}+\rho_{t,g,k}^{2}z^{-2}\right).
\label{eq:token_conditioned_denominator}
\end{equation}
\begin{proposition}
\label{Proposition2}
 Assume that for all $t, g, \ell, k$, we have $|a_{t,g,\ell}| \leq 1-\varepsilon_{\rho}$, $0 < \rho_{t,g,k} \leq 1-\varepsilon_{\rho}$, and $\theta_{t,g,k} \in [0, \pi]$. Then every frozen local denominator $Q_{t,g}$ has all poles strictly inside the unit disk. Hence, each local grouped denominator is Schur stable (see Section \ref{supp:pole_stability_proofs}).
\end{proposition}
\vspace{-4mm}
\paragraph{Low-rank Numerator:}  While the denominator defines the system's poles, a complete transfer function requires a numerator to dictate how past inputs excite these dynamics. Freezing the time-varying coefficients at step $t$ yields a local transfer function $H_t(z) = D + \frac{P_{t,sc}(z^{-1})}{Q_t(z^{-1})}$, which explicitly isolates the zero-lag direct path $D$ from the strictly causal numerator $P_{t,sc}$. We preserve $D$ and construct the causal branch using our token-conditioned denominator alongside a lightweight low-rank numerator.

For a multidimensional feature, a general strictly causal numerator at step $t$ takes the form $P_{t,sc}(z^{-1}) = \sum_{i=1}^{r} B_{t,i} z^{-i}$ with $B_{t,i} \in \mathbb{R}^{E \times E}$, inducing a time-domain driving signal $\eta_t = \sum_{i=1}^{r} B_{t,i} x_{t-i}$. Since constructing the driving signal with dense matrices $B_{t,i}$ is computationally expensive, we approximate this numerator via a low-rank factorization.

Tokens are first projected into a compact latent space $\phi_t = V^\top x_t \in \mathbb{R}^{r_f}$ ($V \in \mathbb{R}^{E \times r_f}$, $r_f < E$), and the past $r$ steps are assembled into a zero-padded causal window $\Phi_t = [\phi_{t-1}, \ldots, \phi_{t-r}] \in \mathbb{R}^{r_f \times r}$. From the current token $x_t$, we then predict token-dependent numerator coefficients and gating weights:
\begingroup
\setlength{\abovedisplayskip}{2pt}
\setlength{\abovedisplayshortskip}{2pt}
\setlength{\belowdisplayshortskip}{1pt}
\setlength{\belowdisplayshortskip}{1pt}
\begin{equation}
\alpha_t = W_\alpha x_t \in \mathbb{R}^{r},
\qquad
\gamma_t = \sigma(W_\gamma x_t) \in \mathbb{R}^{r_f},
\label{eq:numerator_coeffs_gates}
\end{equation}
\endgroup
where trainable parameters $W_\alpha \in \mathbb{R}^{r \times E}$ and $W_\gamma \in \mathbb{R}^{r_f \times E}$.

The mixing coefficients $\alpha_t$ blend the causal window into a low-rank latent mixture using $\psi_t = \Phi_t \alpha_t = \sum_{i=1}^{r} \alpha_{t,i}\phi_{t-i} \in \mathbb{R}^{r_f}$. Gating and up-projecting this mixture forms the final multidimensional driving signal such that:
\begingroup
\setlength{\abovedisplayskip}{1pt}
\setlength{\abovedisplayshortskip}{1pt}
\setlength{\belowdisplayshortskip}{2pt}
\setlength{\belowdisplayshortskip}{2pt}
\begin{equation}
\eta_t = U(\gamma_t \odot \psi_t) \in \mathbb{R}^{E},
\qquad
U \in \mathbb{R}^{E \times r_f},
\label{eq:low_rank_numerator_output}
\end{equation}
\endgroup
where $\odot$ denotes elementwise multiplication. Equivalently, the induced low-rank numerator can be written as
\begin{equation}
P_{t,\mathrm{sc}}^{\mathrm{LR}}(z^{-1})
=
U \operatorname{Diag}(\gamma_t)
\left(\sum_{i=1}^{r} \alpha_{t,i} z^{-i}\right)
V^\top
=
\sum_{i=1}^{r} \alpha_{t,i}\, U \operatorname{Diag}(\gamma_t) V^\top z^{-i}.
\label{eq:low_rank_strictly_causal_numerator}
\end{equation}

Hence, the dense matrices $B_{t,i}$ are represented implicitly through the rank-$r_f$ factorization such that $B_{t,i} = \alpha_{t,i}\, U \operatorname{Diag}(\gamma_t) V^\top.$ Finally, we partition the driving signal into $G$ groups, $\eta_t = [\eta_t^{(1)}; \ldots; \eta_t^{(G)}]$, ensuring each group receives the portion corresponding to its denominator $Q_{t,g}$. Given the expanded token-conditioned denominator $Q_{t,g}(z^{-1}) = 1 + \sum_{i=1}^{r} q_{t,g,i} z^{-i}$, the grouped recurrent branch is computed directly as:
\begin{equation}
y_t^{(g)} = \eta_t^{(g)} - \sum_{i=1}^{r} q_{t,g,i} y_{t-i}^{(g)}.
\label{eq:grouped_recurrent_branch}
\end{equation}
Concatenating these groups yields the full recurrent state $y_t \in \mathbb{R}^{E}$. Lastly, incorporating a learned diagonal direct path $D \in \mathbb{R}^{E \times E}$ (as in Vision Mamba), yields the output as $o_t = y_t + D x_t$. This preserves local token information and stabilizes optimization when long-range recurrence is unnecessary. Ultimately, this explicitly isolates architectural roles: the denominator governs memory via stable poles, the low-rank numerator dictates state excitation, and the direct path preserves instantaneous context.

\vspace{-3mm}
\paragraph{Computational efficiency:} Given a single scan route, our proposed method has a dominant per-token-channel inference cost of $\mathcal{O}(2r + 3r_f)$. In contrast, a standard selective SSM incurs an $\mathcal{O}(7N)$ cost in addition to the step-size projection. Thus, by keeping $r$ and $r_f$ small, the proposed operator achieves computational efficiency while maintaining linear scaling with respect to sequence length $M$ and channel dimension $E$ (see Section \ref{Proof: FLOPS}).
\vspace{-3mm}
\paragraph{TCP-SSM Model Training:} We train our token-conditioned pole model via feature-level distillation using the pre-trained original SSM model as a teacher. Both share identical backbones, differing only where the student introduces our proposed recurrent operator. The trainable parameters of the TCP-SSM operator include the base denominator pole parameters $\hat{\rho}_{g,k}$, $\hat{\theta}_{g,k}$, $\hat{\rho}^{R}_{g,\ell}$, and $\hat{s}_{g,\ell}$, the token-modulation heads $W_\rho$, $b_\rho$, $W_\theta$, and $b_\theta$, the low-rank numerator parameters $U$, $V$, $W_\alpha$, and $W_\gamma$, and the diagonal direct path $D$. The denominator coefficients $q_{t,g,i}$, numerator coefficients $\alpha_t$, and gates $\gamma_t$ are generated from these learned parameters and the current token $x_t$. Letting $o_{b,t}^{(\ell)}$ and $\tilde{o}_{b,t}^{(\ell)}$ denote the student and teacher outputs at layer $\ell$, batch $b$, and token $t$, we compute an $L_1$ loss across all intermediate feature maps:
\begingroup
\setlength{\abovedisplayskip}{0pt}
\setlength{\abovedisplayshortskip}{0pt}
\setlength{\belowdisplayshortskip}{2pt}
\setlength{\belowdisplayshortskip}{2pt}
\begin{equation}
\mathcal{L}_{\mathrm{distill}} =
\frac{1}{L_{\mathrm{blk}}}
\sum_{\ell=1}^{L_{\mathrm{blk}}}
\frac{1}{B M_\ell E_\ell}
\sum_{b=1}^{B}
\sum_{t=1}^{M_\ell}
\left\| o_{b,t}^{(\ell)} - \operatorname{sg}\!\left(\tilde{o}_{b,t}^{(\ell)}\right) \right\|_1.
\label{eq:distill_loss}
\end{equation}
\endgroup
Here $L_{\mathrm{blk}}$ is the number of distilled layers, $M_\ell$ the token count, $E_\ell$ the channel dimension, and $\operatorname{sg}(\cdot)$ the stop-gradient. The total loss $\mathcal{L} = \mathcal{L}_{\mathrm{task}} + \lambda_{\mathrm{distill}} \mathcal{L}_{\mathrm{distill}}$ leverages this intermediate guidance to encourage our efficient operator to emulate the teacher's representations while jointly optimizing the final downstream task objective $\mathcal{L}_{\mathrm{task}}$.
\vspace{-4mm}
\section{Experiments}
\label{Main:Experiments}
\vspace{-3mm}
\subsection{Image Classification}
\vspace{-2mm}
\paragraph{Settings:} We evaluate on the ImageNet-1K dataset \cite{deng2009imagenet} and report single-crop top-1 and top-5 validation accuracies. For fair comparison, we follow the DeiT \cite{pmlr-v139-touvron21a} training recipe with AdamW optimization, cosine learning-rate decay, and standard augmentations including random resized cropping, horizontal flipping, label smoothing, and MixUp \cite{zhang2017mixup}. All experiments are performed on six RTX PRO 6000 Blackwell GPUs. Detailed training hyperparameters are provided in Table \ref{tab:supp_cls_training_hparams}, and TCP-SSM configurations are summarized in Table \ref{supptab:tcp_configs}.

\begin{table}[t]
\centering
\footnotesize
\setlength{\tabcolsep}{2.0pt}
\setlength{\extrarowheight}{3pt}
\renewcommand{\arraystretch}{0.8}
\caption{Comparison on ImageNet-1K. SSM FLOPs are reported for models with an SSM branch.}
\label{tab:imagenet_main}
\begin{tabularx}{0.94\columnwidth}{|
Y|
C{0.13\columnwidth}|
C{0.07\columnwidth}|
C{0.09\columnwidth}|
C{0.15\columnwidth}|
C{0.12\columnwidth}|
C{0.12\columnwidth}|
}
\hline
\textbf{Method} &
\textbf{Token mixing} &
\textbf{Input} &
\textbf{\#Param.}$\boldsymbol{\downarrow}$ &
\textbf{SSM FLOPs} $\boldsymbol{\downarrow}$ &
\textbf{Top-1}\textbf{ (\%)} $\boldsymbol{\uparrow}$ &
\textbf{Top-5}\textbf{ (\%)} $\boldsymbol{\uparrow}$ \\
\hline
ViT-B/16 \cite{dosovitskiy2020image}  & Attention & $384^2$ & 86M  & --    & 77.9 & 95.3 \\
Swin-T \cite{liu2021swin}  & Attention & $224^2$ & 29M  & --    & 81.3 & 95.6 \\
\hline
\hline
QuadMamba-Li \cite{xie2024quadmamba}         & SSM             & $224^2$ & 6.5M  & 318M  & 74.2 & 92.1 \\
Vim-Ti-F \cite{zhang2024vim}         & SSM             & $224^2$ & 7.1M  & 497M & 76.7 & 93.6 \\
LocalViM-T \cite{huang2024localmamba}           & SSM             & $224^2$ & 8M  & 825M  & 76.3 & 93.1 \\
LocalViM-S  \cite{huang2024localmamba}      & SSM             & $224^2$ & 27.8M & 1938.8 & 81.8 & 95.6 \\
S4ND-ViT-B \cite{nguyen2022s4nd}   & SSM + Attention & $224^2$ & 89M   & 414M  & 80.4 & --   \\
PlainMamba-L2 \cite{yang2024plainmamba}   & SSM             & $224^2$ & 25.7M & 1459M & 81.6 & 95.6 \\
Mamba-ND \cite{li2024mamba} & SSM             & $224^2$ & 24M   & 1308M & 81.7 & --   \\
VMamba-T \cite{liu2024vmamba}        & SSM             & $224^2$ & 30.2M & 253M  & 82.5 & 95.9 \\
\hline

Vim-T \cite{zhu2024vision} & SSM & $224^2$ & 7.1M & 497.5M & 76.1 & 93.0 \\
\oursmethod{TCP-Vim-T1} &
\ourscell{SSM} &
\ourscell{$224^2$} &
\ourscell{7.5M} &
\ourscell{\makecell[c]{\textbf{295.3M} \scriptsize($\downarrow 40.6\%$)}} &
\ourscell{{76.0}} &
\ourscell{\textbf{93.1}} \\
\oursmethod{TCP-Vim-T2} &
\ourscell{SSM} &
\ourscell{$224^2$} &
\ourscell{8.1M} &
\ourscell{\makecell[c]{\textbf{459.5M} \scriptsize($\downarrow 7.6\%$)}} &
\ourscell{\textbf{76.6}} &
\ourscell{\textbf{93.3}} \\
\hline

EVM-T \cite{lee2025efficientvim} & SSM & $224^2$ & 6.5M  & 109.3M & 76.3 & 93.1 \\
EVM-B \cite{lee2025efficientvim} & SSM & $224^2$ & 33.7M & 233M & 81.8 & 95.6 \\
\oursmethod{TCP-EVM-T1} &
\ourscell{SSM} &
\ourscell{$224^2$} &
\ourscell{6.4M} &
\ourscell{\makecell[c]{\textbf{62.9M} \scriptsize($\downarrow 42.5\%$)}} &
\ourscell{{76.3}} &
\ourscell{{93.0}} \\
\oursmethod{TCP-EVM-B3} &
\ourscell{SSM} &
\ourscell{$224^2$} &
\ourscell{33.5M} &
\ourscell{\makecell[c]{\textbf{129.9M} \scriptsize($\downarrow 44.2\%$)}} &
\ourscell{\textbf{82.0}} &
\ourscell{\textbf{95.8}} \\
\oursmethod{TCP-EVM-B5 } &
\ourscell{SSM} &
\ourscell{$224^2$} &
\ourscell{33.5M} &
\ourscell{\makecell[c]{\textbf{231.4M} \scriptsize($\downarrow 0.7\%$)}} &
\ourscell{\textbf{82.6}} &
\ourscell{\textbf{96.0}} \\
\hline
\end{tabularx}
\vspace{-5mm}
\end{table}
To study the efficiency trade-off, we replace all 24 SSM layers in Vision Mamba-Tiny \cite{zhu2024vision} with TCP-SSM and train two variants, TCP-Vim-T1 and TCP-Vim-T2, with TCP-Vim-T2 using a higher SSM FLOP budget. Both models are trained for 600 epochs with a learning rate  of $1 \times 10^{-4}$. To evaluate generality, we apply TCP-SSM to EfficientVMamba \cite{pei2025efficientvmamba}, an ideal testbed since it modifies the scan pattern rather than the recurrence. We replace the SSM modules in the first two stages and keep the remaining architecture unchanged. We explore three variants. TCP-EVM-T1 trains for 700 epochs, while TCP-EVM-B1 and TCP-EVM-B2 train for 250 epochs. All use a batch size of 256 and a $1 \times 10^{-4}$ base learning rate.
\paragraph{Results:}

Table \ref{tab:imagenet_main} evaluates the proposed TCP design on Vision Mamba and EfficientVMamba, demonstrating its generality across distinct visual SSM backbones. On the Vim backbone, TCP-Vim-T1 reduces SSM FLOPs by 40.6\% (497.5M to 295.3M) while maintaining comparable Top-1 (76.0\%) and Top-5 (93.1\%) accuracy. TCP-Vim-T2 lowers the SSM cost to 459.5M and improves baseline accuracy to 76.6\% Top-1 and 93.3\% Top-5. EfficientVMamba shows a similar trend. TCP-EVM-T1 cuts SSM FLOPs by 42.5\% (109.3M to 62.9M) while preserving the 76.3\% baseline Top-1 accuracy. For EfficientVMamba-Base, TCP-EVM-B3 achieves a 44.2\% FLOP reduction (233.0M to 129.9M) and improves accuracy to 82.0\% Top-1 and 95.8\% Top-5. When matched to the original compute budget, TCP-EVM-B5 reaches 82.6\% Top-1 and 96.0\% Top-5 accuracy using 231.4M SSM FLOPs. Ultimately, these results confirm that TCP's benefits are architectural rather than backbone-specific, consistently maintaining strong performance with a graceful accuracy - FLOPs trade-off.
\vspace{-3mm}
\subsection{Semantic Segmentation} 
\vspace{-2mm}
\paragraph{Settings:} To evaluate semantic segmentation performance, we conduct experiments on the ADE20K dataset \cite{zhou2017scene} using UPerNet \cite{xiao2018unified}. The backbone is initialized with ImageNet-pretrained weights and followed by a UPerHead for dense prediction. We optimize the model using AdamW with a learning rate of $1 \times 10^{-4}$ and a weight decay of 0.01, training for $180K$ iterations with total batch size of 24.
\vspace{-3mm}
\paragraph{Results:} 
\begin{table}[t]
\centering
\footnotesize
\setlength{\tabcolsep}{1pt}
\setlength{\extrarowheight}{2pt}
\renewcommand{\arraystretch}{0.82}
\caption{Comparison on ADE20K semantic segmentation. * denotes our produced results.}
\label{tab:ade20k_main}
\begin{tabular}{|
>{\centering\arraybackslash}p{0.20\columnwidth}|
>{\centering\arraybackslash}p{0.14\columnwidth}|
>{\centering\arraybackslash}p{0.08\columnwidth}|
>{\centering\arraybackslash}p{0.10\columnwidth}|
>{\centering\arraybackslash}p{0.15\columnwidth}|
>{\centering\arraybackslash}p{0.13\columnwidth}|}
\hline
\textbf{Method} &
\textbf{Token mixing} &
\textbf{Input} &
\textbf{\#Param.}$\boldsymbol{\downarrow}$ &
\textbf{SSM FLOPs} $\boldsymbol{\downarrow}$ &
\textbf{mIoU (\%)} $\boldsymbol{\uparrow}$ \\
\hline
SegFormer-B0 \cite{xie2021segformer} & Attention & $512^2$ & 3.7M & -- & 37.4 \\
DeiT-T \cite{touvron2021training} & Attention & $512^2$ & 11M & -- & 39.2 \\
\hline
\hline
Octopus-T \cite{mahatha2026octopus} & SSM & $512^2$ & 63.3M & 0.5G & 37.9 \\
Octopus-B \cite{mahatha2026octopus} & SSM & $512^2$ & 85.8M & 1.5G & 39.0 \\
EfficientVMamba-T \cite{lee2025efficientvim} & SSM & $512^2$ & 14M & 0.6G & 38.9 \\
Vim-T + DyVM \cite{wu2025dynamic} & SSM & $512^2$ & 12.5M & 1.8G & 40.1 \\
Vim-Ti-F \cite{zhang2024vim} & SSM & $512^2$ & 34M & 2.58G & 40.4 \\
\hline
Vim-T$^\ast$ \cite{{zhu2024vision}}  & SSM & $512^2$ & 12.3M & 2.58G & 40.6 \\
\oursmethod{TCP-Vim-T1} &
\ourscell{SSM} &
\ourscell{$512^2$} &
\ourscell{12.7M} &
\ourscell{\makecell[c]{\textbf{1.53G} \scriptsize($\downarrow 40.6\%$)}} &
\ourscell{\textbf{41.1}} \\
\hline
\end{tabular}
\vspace{-5mm}
\end{table}
\begin{table}[t]
\centering
\footnotesize
\setlength{\tabcolsep}{1.0pt}
\setlength{\extrarowheight}{1pt}
\renewcommand{\arraystretch}{0.85}
\caption{Object detection and instance segmentation on COCO dataset.}
\label{tab:coco_det_seg}
\resizebox{\columnwidth}{!}{%
\begin{tabular}{|
>{\centering\arraybackslash}p{0.12\columnwidth}|
>{\centering\arraybackslash}p{0.07\columnwidth}|
>{\centering\arraybackslash}p{0.14\columnwidth}|
>{\centering\arraybackslash}p{0.08\columnwidth}|
>{\centering\arraybackslash}p{0.08\columnwidth}|
>{\centering\arraybackslash}p{0.08\columnwidth}|
>{\centering\arraybackslash}p{0.08\columnwidth}|
>{\centering\arraybackslash}p{0.08\columnwidth}|
>{\centering\arraybackslash}p{0.08\columnwidth}|}
\hline
\textbf{Method} &
\textbf{\#Param.} &
\textbf{SSM FLOPs} $\boldsymbol{\downarrow}$ &
$\mathbf{AP}^{box}$ &
$\mathbf{AP}^{box}_{50}$ &
$\mathbf{AP}^{box}_{75}$ &
$\mathbf{AP}^{mask}$ &
$\mathbf{AP}^{mask}_{50}$ &
$\mathbf{AP}^{mask}_{75}$ \\[2pt]
\hline
Vim-T \cite{zhu2024vision} &
60.0M &
10.34G &
45.7 & 63.9 & 49.6 &
39.2 & {60.9} & 41.7 \\
\rowcolor{gray!25}
\textbf{TCP-Vim-T1} &
{60.4M} &
\textbf{6.14G} {\scriptsize($\downarrow 40.6\%$)} &
\textbf{47.8} & \textbf{64.2} & \textbf{52.6} &
\textbf{39.3} & 60.6 & \textbf{43.1} \\
\hline
\end{tabular}%
}
\vspace{-5mm}
\end{table}

Table \ref{tab:ade20k_main} presents semantic segmentation results on the ADE20K validation set, demonstrating the transferability of our proposed TCP design. Since a pretrained Vim-T segmentation checkpoint is publicly unavailable, we train this baseline under our experimental setup to ensure a fair comparison. Building upon this, TCP-Vim-T1 improves the baseline mIoU from 40.6\% to 41.1\% while simultaneously reducing SSM FLOPs by 40.6\% (from 2.58G to 1.53G). This achievement is especially significant as dense prediction dictates much stricter demands on spatial granularity than standard image classification. The concurrent improvement in performance and reduction in computational overhead confirm that our token-conditioned pole mechanism does not compromise representational capacity for efficiency. 
\vspace{-3mm}
\subsection{Object Detection and Instance Segmentation}
\vspace{-2mm}
\paragraph{Settings:} We evaluate TCP-SSM on COCO 2017 \cite{lin2014microsoft} for object detection and instance segmentation within a ViTDet-style framework using Simple Feature Pyramid \cite{li2022exploring} and Cascade Mask R-CNN \cite{cai2019cascade} heads. We replace only the Vim backbone with TCP-SSM and report box AP and mask AP. We provide detailed settings in Section \ref{supp:Implementation}.
\vspace{-3mm}
\paragraph{Results:} Table \ref{tab:coco_det_seg} presents object detection and instance segmentation results on COCO, demonstrating that the proposed TCP design transfers effectively to detection-based dense prediction tasks. TCP-Vim-T1 maintains enhanced downstream performance to the Vim-T baseline while reducing SSM FLOPs by 40.6\% (10.34G to 6.14G).
\vspace{-3mm}
\subsection{Ablation Study}
\vspace{-2mm}
\paragraph{Performance vs. Efficiency:}

\begin{wrapfigure}{r}{0.6\textwidth}
    \centering
    \includegraphics[width=0.55\textwidth]{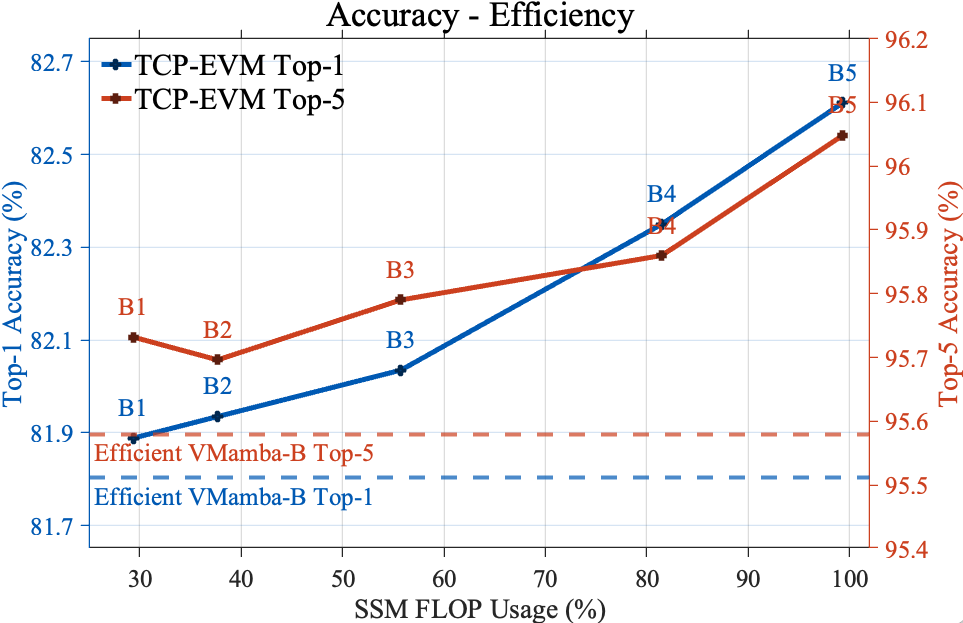}
    \vspace{-0.4em}
    \caption{Accuracy-computation trade-off on EVM-B.}
    \label{fig:ablation_study_1}
    \vspace{-2mm}
\end{wrapfigure}

Figure \ref{fig:ablation_study_1} illustrates the accuracy-computation trade-off for EfficientVMamba-Base when the original SSM modules are replaced by our proposed TCP design across various configurations. This comparison evaluates how recognition performance scales as recurrent computation is progressively reduced while the backbone architecture remains fixed. While the baseline yields 81.8\% Top-1 and 95.6\% Top-5 accuracy (233M SSM FLOPs), our TCP-EVM-B5 variant elevates this performance to 82.6\% Top-1 and 96.0\% Top-5 at a slightly reduced computational cost of 0.7\%. The efficiency gains become even more pronounced at higher reduction levels. TCP-EVM-B4 delivers 82.4\% Top-1 with an 18.5\% FLOP reduction, and TCP-EVM-B3 retains 82.0\% Top-1 with a 44.2\% drop in FLOPs. Most notably, TCP-EVM-B2 and TCP-EVM-B1 aggressively reduce SSM computation by 62.3\% and 70.6\%, yet still defend a strong 81.9\% Top-1 accuracy. Overall, these results demonstrate a highly favorable efficiency trend across a broad range of operating points, confirming that our model successfully preserves strong accuracy despite substantial reductions in recurrent cost. (Detailed configurations in Table \ref{supptab:tcp_configs}).
\vspace{-3mm}
\paragraph{Impact of Distillation Loss:}
Table \ref{tab:distill_ablation} investigates the impact of distillation loss in TCP-Vim-T1. In the absence of distillation, performance degrades (75.3\% Top-1, 92.7\% Top-5), indicating that TCP replacement alone struggles to recover the full accuracy of the Vim-T baseline. Adding distillation consistently improves performance across the tested weights, demonstrating that teacher guidance effectively helps the proposed recurrent module preserve the original backbone behavior. Specifically, $\lambda_{\mathrm{distill}}=6\times10^{-2}$ yields the strongest results (76.0\% Top-1, 93.0\% Top-5). This closes the performance gap while retaining the reduced SSM computation of TCP-Vim-T1.
\vspace{-3mm}
\paragraph{Effect of Token-Conditioned Pole Modulation:}
\label{Ablation_Supp}

\begin{table}[t]
\centering
\makebox[\columnwidth][c]{%
\begin{minipage}[t]{0.47\columnwidth}
\centering
\scriptsize
\setlength{\tabcolsep}{1.3pt}
\setlength{\extrarowheight}{2pt}
\renewcommand{\arraystretch}{1}
\captionof{table}{Ablation Study on $\mathcal{L}_{\mathrm{distill}}$.}
\label{tab:distill_ablation}
\resizebox{\linewidth}{!}{%
\begin{tabular}{|
>{\centering\arraybackslash}p{0.23\linewidth}|
>{\centering\arraybackslash}p{0.11\linewidth}|
>{\centering\arraybackslash}p{0.15\linewidth}|
>{\centering\arraybackslash}p{0.1\linewidth}|
>{\centering\arraybackslash}p{0.1\linewidth}|}
\hline
\textbf{Method} &
\makecell[c]{Distill.} &
\makecell[c]{\boldmath$\lambda_{\mathrm{distill}}$} &
\textbf{Top-1} &
\textbf{Top-5} \\
\hline
Vim-T & -- & -- & 76.1 & 93.0 \\
\hline
TCP-Vim-T1 & $\times$ & -- & 75.3 & 92.7 \\
TCP-Vim-T1 & $\checkmark$ & $10^{-1}$ & 75.8 & 92.9 \\
TCP-Vim-T1 & $\checkmark$ & $2 \times 10^{-2}$ & 75.6 & 92.8 \\
\rowcolor{gray!25}
\textbf{TCP-Vim-T1} & \textbf{$\checkmark$} & \textbf{$6 \times 10^{-2}$} & \textbf{76.0} & \textbf{93.0} \\
\hline
\end{tabular}
}
\end{minipage}%
\hspace{2pt}%
\begin{minipage}[t]{0.53\columnwidth}
\centering
\footnotesize
\setlength{\tabcolsep}{1pt}
\setlength{\extrarowheight}{1pt}
\renewcommand{\arraystretch}{0.85}
\captionof{table}{Ablation study on pole selectivity.}
\label{tab:pole_selectivity_ablation}
\resizebox{\linewidth}{!}{%
\begin{tabular}{|
>{\centering\arraybackslash}p{0.22\linewidth}|
>{\centering\arraybackslash}p{0.14\linewidth}|
>{\centering\arraybackslash}p{0.14\linewidth}|
>{\centering\arraybackslash}p{0.19\linewidth}|
>{\centering\arraybackslash}p{0.10\linewidth}|
>{\centering\arraybackslash}p{0.10\linewidth}|}
\hline
\textbf{Method} &
\textbf{TCP} &
\textbf{\#Param.} &
\textbf{SSM FLOPs} &
\textbf{Top-1} &
\textbf{Top-5} \\
\hline
Vim-T & -- & 7.1M & 497.5M & 76.1 & 93.0 \\
\hline
TCP-Vim-T1 & $\times$ & 7.4M & 280.6M & 75.4 & 92.7 \\
TCP-Vim-T1 & \makecell[c]{$\checkmark$\\G-spec.} & 7.9M & 375.3M & 76.1 & 93.1 \\
\rowcolor{gray!25}
\textbf{TCP-Vim-T1} &
\makecell[c]{\textbf{$\checkmark$}\\\textbf{Shared}} &
\textbf{7.5M} &
\textbf{295.3M} &
76.0 &
93.0 \\
\hline
\end{tabular}
}
\end{minipage}%
}
\vspace{-6mm}
\end{table}
Table \ref{tab:pole_selectivity_ablation} evaluates token-conditioned poles ($G=12$), which enable recurrent dynamics to adapt to local image content. With a fixed pole structure, TCP-Vim-T1 reaches 75.4\% Top-1 accuracy, suggesting that static poles limit the flexibility of the recurrent module. Introducing token-conditioning improves performance, though the group-specific variant increases SSM FLOPs to 375.3M (76.1\% Top-1, 93.1\% Top-5). The shared-selectivity design provides a better balance, achieving comparable accuracy of 76.0\% Top-1 and 93.0\% Top-5 with a lower cost of 295.3M SSM FLOPs. This shows that sharing the modulation across groups preserves most of the benefit of token-conditioned poles while reducing the recurrent computation (Ablation study of $G$ in Table \ref{supp:Abl_Group_Number_Table}).
\vspace{-4mm}
\section{Conclusion}
\label{Main: Conclude}
\vspace{-3mm}
We present an efficient visual state space modeling framework that uses token-conditioned poles to explicitly parameterize the recurrence dynamics of Vision Mamba-style models through stable real and complex pole components. By separating monotone memory decay from oscillatory frequency responses, our formulation provides a more interpretable and controllable alternative to conventional selective scans. Token-conditioned modulation enables stable, content-adaptive spatial memory, while grouped pole sharing and a lightweight low-rank input pathway preserve linear-time scan complexity. Across classification, segmentation, and object detection, the proposed approach improves the accuracy-efficiency trade-off of compact visual SSMs, reducing SSM computation by up to 44\% in Vision Mamba and 70\% in EfficientVMamba-Base while maintaining or improving accuracy.
\vspace{-4mm}
\paragraph{Limitations and Future Work:} Our method introduces additional parameters, such as pole order, group size, and low-rank dimension, which may require deployment-specific tuning. While the pole parameterization ensures stable local recurrence dynamics, it does not directly address other sources of inefficiency, such as scan ordering and memory access patterns. Future work can explore native two-dimensional pole-structured recurrences, adaptive scan, and hardware-aware implementations.

\bibliographystyle{plainnat}
\bibliography{references}
\clearpage
\appendix
 \end{document}